\documentclass[lettersize,journal]{IEEEtran}
\usepackage{amsmath,amsfonts}
\usepackage{algorithmic}
\usepackage{algorithm}
\usepackage{array}
\usepackage[caption=false,font=normalsize,labelfont=sf,textfont=sf]{subfig}
\usepackage{textcomp}
\usepackage{stfloats}
\usepackage{url}
\usepackage{verbatim}
\usepackage{graphicx}
\usepackage{cite}
\usepackage{subcaption}
\usepackage{bm}
\newcommand{\mbf}[1]{\mathbf{#1}}
\begin{document}
	
	\title{A Dynamic Mode Decomposition Approach to\\ Morphological Component Analysis}
	
	\author{Owen T. Huber, Raghu G. Raj, Tianyu Chen, Zacharie I. Idriss
		\thanks{DISTRIBUTION STATEMENT A. Approved for public release; distribution is unlimited.}
		\thanks{This research is supported by the Office of Naval Research (ONR).}
		\thanks{Owen T. Huber, Raghu G. Raj, and Tianyu Chen are with the Radar Division, U.S. Naval Research Laboratory. Zacharie I. Idriss was with the Radar Division during the writing of this manuscript.}
		}

	\maketitle
	
	\begin{abstract}
		This paper introduces a novel methodology of adapting the representation of videos based on the dynamics of their scene content variation. In particular, we demonstrate how the clustering of dynamic mode decomposition eigenvalues can be leveraged to learn an adaptive video representation for separating structurally distinct morphologies of a video. We extend the morphological component analysis (MCA) algorithm, which uses multiple predefined incoherent dictionaries and a sparsity prior to separate distinct sources in signals, by introducing our novel eigenspace clustering technique to obtain data-driven MCA dictionaries, which we call dynamic morphological component analysis (DMCA). After deriving our novel algorithm, we offer a motivational example of DMCA applied to a still image, then demonstrate DMCA's effectiveness in denoising applications on videos from the Adobe 240fps dataset. Afterwards, we provide an example of DMCA enhancing the signal-to-noise ratio of a faint target summed with a sea state, and conclude the paper by applying DMCA to separate a bicycle from wind clutter in inverse synthetic aperture radar images. 
	\end{abstract}
	
	
	\section*{Nomenclature}
	\addcontentsline{toc}{section}{Nomenclature}
	\begin{IEEEdescription}[\IEEEusemathlabelsep\IEEEsetlabelwidth{$V_1,V_2,V_3$}]
		\item[$A$] Uppercase characters are matrices.
		\item[$A_{:, j}$] $j^{th}$ column of $A$.
		\item[$A_{:, [a, b]}$] Matrix with columns $A_{:, j}$ for all $j \in [a, b]$.
		\item[$A_{i, j}$] Scalar at index $(i, j)$ of $A$.
		\item[$A^*$] Conjugate transpose of $A$.
		\item[col$(A)$] Column space of $A$.
		\item[$\mathbf{a}$] Boldface lowercase characters are vectors.
		\item[$a_i$] Scalar at index $i$ of the vector $\mathbf{a}$.
		\item[$x_{(p)}$] Element of a one-dimensional sequence at index $p$
		\item[$x_{(p, q)}$] Element of a two-dimensional sequence at index $p, q$. 
	\end{IEEEdescription}

	\section{Introduction}
	Separating a video into its distinct, or `natural,' components such as cartoons, coherent temporal structures, clutter, or noise, is a crucial step in video processing, and often determines whether a video offers any value for understanding the scene it captures. Key information for this problem is contained in the dynamics of ordered frames (where a frame is one of the still images that comprise a video), but the high dimensionality of the data and lack of governing equations make the problem of decomposing a video into its source signals a daunting task. 
	
	The numerous approaches that have been developed for video representation in recent decades, with analogous image schemes, can be categorized in several different ways. Mathematical approaches involve creating a correspondence between function spaces (such as $L^2{\left(\mathbb{R}^2\right)}$ \cite{Kreyszig2007}) and videos, and using basis sets, frames, or, in general, dictionaries for individual image representation. Examples of such approaches include the use of Fourier \cite{oppenheim1975digital}, wavelets \cite{mallat08} and space-frequency dictionaries \cite{Grochenig2001foundations}, many of which are associated with computationally efficient transforms (such as the Fast Fourier Transform (FFT) \cite{oppenheim1975digital}) to create a mapping between image and representation spaces. Statistical approaches involve analyzing image patches, which may be sampled from a video, to synthesize a dictionary for image representation. Examples of this include the K-means Singular Value Decomposition (K-SVD) \cite{ksvd06}, union of orthobases \cite{Dictionaries}, and principle component analysis (PCA) \cite{pmlr-v9-jenatton10a} approaches. More recently, generative machine learning (ML) methods have been developed for dictionary creation that portend to be one of the most powerful techniques for image and video representation \cite{pmlr-v9-jenatton10a}.
	
	It is important to note that dictionary creation or optimization methods also depend on the intended video representation applications. For example, video representation schemes, under any of the above paradigms, can be tailored for regression \cite{sparse_dictionaries} or data classification \cite{wright09, mckaymongaraj17} applications. Another application, of particular interest in this paper, is that of creating data-fidelity-based representation schemes that simultaneously model temporally or structurally distinct components of a video (that need not be spatially disjoint). 
	
	This paper introduces a new methodology of dictionary creation wherein the evolution of a data-driven dynamical systems model is used to inform the creation of adaptive dictionaries. These dictionaries are suitable for morphological component analysis (MCA) applications, which involves utilizing a set of mutually incoherent dictionaries that capture structurally distinct features in signals via the solution of an $\ell_1$ linear inverse problem \cite{MCA}. The particular dynamical systems modeling technique that we employ is dynamic mode decomposition (DMD), which offers a computationally efficient means of extracting coherent structures from time-series data and modeling how they evolve over time. Since its inception, a multitude of variants of DMD have been developed for specialized applications, including multiresolution DMD, physics-informed DMD, and DMD with control \cite{COLBROOK2024127}. In DMCA, we use the fundamental DMD algorithm which will be explained in Section \ref{sec:dmd}.
	
	In later sections of this paper, we construct methods of clustering in DMD eigenspace to inform the separation of DMD modes into distinct dictionaries where they act as atoms for MCA. The separation of modes into background and sparse components by the magnitude of their DMD eigenvalues, with a robust PCA-like matrix separation method, have been used for surveillance video applications in \cite{grosek2014dynamicmodedecompositionrealtime} to remove stationary objects in real-time. We take an alternate approach, using an extended clustering method in DMD eigenspace to inform dictionaries, which are used to separate a single video into $k$ component videos. 
	
	Our work investigates a mathematical approach to video decomposition, which may complement recent deep learning image and video processing algorithms. Rather than competing with deep learning based approaches, we aim to better understand the spectral principles that underlie an MCA-based separation approach as we expect insights from these approaches to offer value in certain contexts. We observe that some current models have begun integrating classical physics frameworks and continuous flow principles into their architectures\cite{kamilov2017plug}\cite{croitoru2023diffusion}. This integration suggests that mathematical and physical understanding continues to have relevance when integrated with data-driven approaches.
	
	Though DMD is a data-driven method, our use of it requires no training data beyond the video to be decomposed and no labeled data. The use of DMD for dictionary creation, and component separation, offers a way to learn video-representation schemes in a robust, efficient, and interpretable way. While our approach is limited in scope compared to end-to-end learning systems, we expect it to offer complementary insights for specific imaging challenges.

	The novel contributions of this paper include 
	\begin{enumerate}
		\item The introduction of DMD for data-driven dictionary generation and the creation of a new algorithm called dynamic morphological component analysis, that constitutes a novel amalgamation of DMD and MCA techniques. 
		\item Demonstration of the performance of DMCA when applied to non-Gaussian noise removal for optical videos in the Adobe 240fps dataset \cite{su2024adobe240fps}, and comparing results with those from Video Block-Matching 4D Filtering (V-BM4D). 
		\item Application of DMCA to increase the signal-to-noise-ratio (SNR) of a target summed with a video of a sea state's height map. 
		\item Demonstration of DMCA's effectivness for separating a bicycle from wind clutter in complex-valued inverse synthetic aperture radar (ISAR) image sequences. 
	\end{enumerate}
	
	We organize the paper as follows: in Section \ref{sec:background}, we introduce both the DMD and MCA methods, along with prior art in DMD eigenvalue clustering. In Section \ref{sec:dmca}, we present the main theatrical contribution on our novel DMCA algorithm. In Section \ref{sec:exp}, we provide the corresponding results where we validate our algorithm on a series of numerical datasets and provide error metrics. In Section \ref{sec:conclusion}, we discuss our results and provide extensions for future work. 
	
	\section{Relevant Background}\label{sec:background}
	
	\subsection{Dynamic mode decomposition}\label{sec:dmd}
	
	Consider a time-series matrix whose columns are measurements on a dynamical system with indexes representing the times at which each measurement is captured. Given sequential measurements, $\mathbf{x}_j \in \mathbb{C}^m$ for $j = 1, \hdots, n$, we define the time-series data matrix, $\mathbf{X}$, as $$\mathbf{X} = \begin{bmatrix}
		\mathbf{x}_1 & \mathbf{x}_2 & \hdots & \mathbf{x}_{n}
	\end{bmatrix}$$ and the two data submatrices, $\mathbf{X}_{:, [1, n -1]}$ and $\mathbf{X}_{:, [2, n]}$, as 
	\begin{align*}
		\mathbf{X}_{:, [1, n -1]} &= \begin{bmatrix}
			\mathbf{x}_1 & \mathbf{x}_2 & \hdots & \mathbf{x}_{n - 1}
		\end{bmatrix}\\
		\mathbf{X}_{:, [2, n]} &= \begin{bmatrix}
			\mathbf{x}_2 & \mathbf{x}_3 & \cdots & \mathbf{x}_n \end{bmatrix}.
	\end{align*}
	For a requested $r \leq n - 1$, DMD seeks to find the spectral decomposition of the rank $r$ best fit operator, $A^{opt}$, that satisfies $\mathbf{X}_{:, [2, n]} \approx A \mathbf{X}_{:, [1, n -1]}$. $A^{opt}$ is rigorously defined as 
	\begin{align}\label{eq:scripts}
		A^{opt} &= \underset{A}{\text{arg }\min}||\mathbf{X}_{:, [2, n]} - A\mathbf{X}_{:, [1, n -1]}||_F\\
		&  \qquad \text{such that rank}(A) = r \nonumber
	\end{align}where the subscript $F$ denotes the Frobenius norm. In the case that no dimensionality reduction is performed, that is when $r = n - 1$, we equivalently state that 
	\begin{equation}\label{eq:pseudoinv}
		A^{opt} = \mathbf{X}_{:, [2, n]} (\mathbf{X}_{:, [1, n -1]})^{\dagger}
	\end{equation} where the superscript $\dagger$ denotes the Moore-Penrose pseudo-inverse. The best fit matrix, $A^{opt}$, then establishes a linear operator that best evolves snapshots over time by the relation $$\mathbf{x}_{j+1} \approx A^{opt} \mathbf{x}_j.$$Finding $A^{opt}$ via standard methods, like computing eigenvalues directly from the pseudo-inverses, would be very computationally expensive. Instead, DMD leverages dimensionality reduction to compute the dominant eigenvectors and eigenvalues of $A^{opt}$, which we call the DMD modes and DMD eigenvalues, without any explicit computations using $A^{opt}$ directly.
	
	In the contexts that DMD is used for, including our own in this paper, $m \gg n$. Since $A^{opt}$ is at most rank $n - 1$ while being of size $m\times m$, it's large size and low rank makes operations involving its multiplication and eigenvalue decomposition computationally expensive. Instead of computing $A^{opt}$, we work on $\tilde{A}^{opt}$, the projection of $A^{opt}$ onto its singular vectors. The algorithm developed by Tu in \cite{H_Tu_2014} and reconstructed in \cite{Brunton_Kutz_2019} to compute the DMD modes and eigenvalues is as follows.

	\begin{enumerate}
		\item Compute the SVD of $\mathbf{X}_{:, [1, n -1]}$: \begin{equation}\label{eq:SVD}
			\mathbf{X}_{:, [1, n -1]} \approx \tilde{U}\tilde{\Sigma}\tilde{V}^*
		\end{equation}for which $^*$ denotes the conjugate transpose of a matrix, $\tilde{U} \in \mathbb{C}^{m \times r}, \tilde{\Sigma} \in \mathbb{C}^{r \times r}$, $\tilde{V} \in \mathbb{C}^{n \times r}$ and $r \leq n - 1$ denotes the number of singular values on the diagonal of $\tilde{\Sigma}$. The columns of $\tilde{U}$ are proper orthogonal decomposition (POD) modes and are unitary, and the columns of $\tilde{U}$ are orthonormal, and the matrix is also unitary. 
		\item At this point, we would be able to calculate $A^{opt}$ via $A^{opt} = \mathbf{X}_{:, [2, n]} \tilde{V}\tilde{\Sigma}^{-1}\tilde{U}^*$. However, since we are only interested in the leading $r$ modes of this matrix, we reduce computations by computing the modes of $A^{opt}$ projected onto the POD modes in $\tilde{U}$ as $$\tilde{A}^{opt} = \tilde{U}^* A^{opt} \tilde{U} = \tilde{U}^* \mathbf{X}_{:, [2, n]}\tilde{V}\tilde{\Sigma}^{-1},$$ which has the same nonzero eigenvalues as $A^{opt}$. In this step, we have computed $\tilde{A}^{opt}$ directly without needing computations involving $A^{opt}$. Note that $\tilde{A}^{opt}$ is a linear model for the POD coefficients $\mathbf{\tilde{x}}$ with $\mathbf{\tilde{x}}_{k+1} = \tilde{A}^{opt}\mathbf{\tilde{x}}_k$ and can be mapped back to the full state by $\mathbf{x} = \tilde{U}\mathbf{\tilde{x}}$.
		\item Next, the DMD eigenvalues of $A^{opt}$, which are the same as the eigenvalues of $\tilde{A}^{opt}$, are computed with
		\begin{equation}\label{eq:SpecD}
			\tilde{A}^{opt}W = W \Lambda. 
		\end{equation} $\Lambda$ is a diagonal matrix whose nonzero values are the eigenvalues, and the columns of $W$ are the eigenvectors of $\tilde{A}^{opt}$.
		\item Finally, we recover the high-dimensional modes, which are stored as the columns of $\Phi$, by multiplying the eigenvectors $W$ by $\tilde{U}$ then $A^{opt}$ as follows
		\begin{align*}
			\Phi &= A^{opt} \tilde{U} W\\
			&= \mathbf{X}_{:, [2, n]}\tilde{V}\tilde{\Sigma}^{-1}\tilde{U}^* \tilde{U}W\\
			\Phi &= \mathbf{X}_{:, [2, n]}\tilde{V}\tilde{\Sigma}^{-1}W
		\end{align*}
		Thus, we avoid computations involving the high-dimensional $\tilde{U}$ and $A^{opt}$. Also, we show that the high-dimensional modes still correspond to the low-dimensional eigenvalues, which are the diagonal elements of $\Lambda$ 
		\begin{align*}
			A^{opt}\Phi &= (\mathbf{X}_{:, [2, n]}\tilde{V}\tilde{\Sigma}^{-1}\tilde{U}^*)( \mathbf{X}_{:, [2, n]}\tilde{V}\tilde{\Sigma}^{-1}W)\\
			&= \mathbf{X}_{:, [2, n]}\tilde{V}\tilde{\Sigma}^{-1}\tilde{A}^{opt}W\\
			&= \mathbf{X}_{:, [2, n]}\tilde{V}\tilde{\Sigma}^{-1}W\Lambda\\
			&= \Phi \Lambda
		\end{align*} 
	\end{enumerate}
	
	In summary, DMD returns a modal decomposition of measurements of dynamical systems for prediction and control. Each mode in the decomposition consists of spatial structures that have the same behavior in time, and whose behaviors can be linearly combined to reconstruct the behavior of the whole system. Every DMD mode is associated with a complex eigenvalue that determines the oscillation frequency and decay or growth rate of that mode. In this way, DMD can be viewed as merging the spatial dimensionality reduction aspects of SVD with the temporal frequency detection capabilities of the fast Fourier transform (FFT).
	
	Though initially developed years earlier, Dynamic Mode Decomposition (DMD) gained significant attention within the dynamical systems community following the work of Rowley, Mezić, and collaborators, who demonstrated its connection to the Koopman operator \cite{ROWLEY_MEZIC_BAGHERI_SCHLATTER_HENNINGSON_2009}.  Specifically, they showed that DMD approximates the infinite dimensional Koopman operator restricted to directly measured observables. Koopman operator theory, introduced by Bernard Koopman in 1931, provides a linear, albeit infinite-dimensional, representation of nonlinear dynamical systems acting on a Hilbert space of observables. Although the Koopman operator's spectral decomposition fully characterizes the underlying nonlinear dynamics, its infinite dimensionality poses computational challenges. By extracting dominant modes from the linear operator mapping observables of a dynamical system to subsequent observables, DMD circumvents the challenges that arise from infinite dimensionality, and its equation-free modeling capabilities have made it ``a workhorse algorithm for the data-driven characterization of high-dimensional systems'' \cite{Brunton_Kutz_2019}.

	DMD has become very popular not only due to its strong relation to Koopman theory, but also due to its flexibility in implementation and numerical stability. Even when the dynamics of a system are unknown, DMD can be applied to measurements of the system to blindly discover the underlying dynamics. Variants of DMD have even been applied to learn the governing equations of systems when only limited measurements of the system are available \cite{doi:10.1073/pnas.1517384113}. 
	
	A key property of DMD that we exploit in our DMCA algorithm, and which provides crucial information on a signal source, is the ability to expand measurements, $\mathbf{x}_t$, taken at some time $t$ into its spectral decomposition via the data-driven approach
	\begin{equation}\label{eq:dmdsd}
		\mathbf{x}_t = \sum_{i = 1}^r \Phi_{:, i} \Lambda_{i, i}^{t - 1} b_i = \Phi \Lambda ^{t -1}\mathbf{b}
	\end{equation}
	where the columns of $\Phi$ are the DMD modes, the diagonal values of $\Lambda$ are the DMD eigenvalues, and the entries of $\mathbf{b}$ are the mode amplitudes. For our purposes, these eigenvalues contain information regarding how dominant a mode is in the reconstruction of a video frame, and how that dominance changes as the frames evolve. We refer to the eigenvalue raised to the power of $t$ as the dynamics of the mode associated with that eigenvalue, and exploit this information in our DMCA algorithm detailed in Section \ref{sec:dmca}.

	\subsection{DMD Eigenvalue clustering}
	
	Methods of clustering in DMD eigenspace have been very effective at separating dynamics in multiscale modeling problems. Multiresolution Dynamic Mode Decomposition (MrDMD), developed by Kutz et al.\cite{kutz2015multiresolutiondynamicmodedecomposition}, integrates DMD and multiresolution analysis to separate a complex system into a hierarchy of its time-scale components. Defining ``slow'' modes as those with corresponding eigenvalues whose distance from the origin are greater than some value, MrDMD recursively removes slow modes from the data to capture increasingly fast modes with greater sensitivity.  This method applies DMD to progressively narrower submatrices of the time-series data, with each iteration halving the temporal width of the preceding submatrix, in a process similar to that used in wavelet analysis \cite{kutz2015multiresolutiondynamicmodedecomposition}. The submatrices used in MrDMD are extremely effective at extracting transient structures in the data that may not be present throughout the entire time-series or are difficult to detect with the dominance of concurrent slower structures.
	
	MrDMD was extended by Dylewsky et al.\cite{Dylewsky_2019} by introducing a \textit{sliding DMD window}, which, in place of the window halving scheme that MrDMD employs, extracts eigenvalues from overlapping windowed subsets of the data matrix. Since DMD's ability to robustly identify a component at a particular time-scale is highly sensitive to window size, and may require window sizes that are not a power of two, this scheme enables the extended MrDMD to more accurately capture dynamics that evolve on different scales temporally and spatially. Using a diagnostic on the spectral bands of the DMD eigenvalues to inform the algorithm on the optimal window length, this method separates the time-scales in the data via k-medians clustering on the DMD eigenvalues. In DMCA, we use overlapping sliding DMD windows toward a similar end.
	
	\subsection{Morphological Component Analysis}
	The \textit{morphology} of a signal can be broadly defined as the type of structure that it exhibits. Two signals with different morphologies will exhibit different structures, such as edges, texture, or smoothness, and may each be sparsely represented by different dictionaries. When these two signals are summed together, we say that the resultant signal exhibits multiple morphologies. 
	
	MCA uses the morphological diversity of features in signals with incoherent dictionaries that each sparsely represent exactly one feature in the data to accomplish a wide range of signal processing tasks. MCA has been shown to be effective in separating the texture from the piecewise smooth component in signals \cite{Starck2005ImageDecomposition}, for inpainting applications \cite{Elad2005MCAInpainting}, and more general blind source separation (BSS) problems where the components have different morphologies, which is what we are interested in \cite{Starck2004RedundantTransforms, Starck2005ImageDecomposition}. Under the assumption that the signal is a linear combination of its source signals, MCA seeks to recover the morphological components from the mixture observed \cite{BSS}, \cite{MCA}, \cite{NRL_2017}. 
	
	To use MCA, the following assumptions must be met: 
	
	\begin{enumerate}
		\item The signal, $\mathbf{s}$, to be separated is linearly reconstructed by $k$ different sources, $\{\mathbf{s}_{(p)}\}_{p \leq k}$, as $\mathbf{s} = \sum_{p \leq k} \mathbf{s}_{(p)}$ for which each $\mathbf{s}_{(p)}$ is a different ``type of signal'', or has a different \textit{morphology}. 
		\item For each component, or source, $\mathbf{s}_{(p)}$, there is a \textit{dictionary}, $\Phi_{(p)}$, such that there are some coefficients, $\bm{\alpha}_{(p)}^{\text{opt}}$, satisfying $\mathbf{s}_{(p)} = \Phi_{(p)} \bm{\alpha}_{(p)}^{\text{opt}}$, and $\bm{\alpha}_{(p)}^{\text{opt}}$ is sparse ($||\bm{\alpha}_{(p)}^{\text{opt}}||_0$ is small). 
		\item For all $q \neq p$, coefficients, $\bm{\alpha}_{(q)}$, satisfying 
		\begin{equation*}
			\text{arg }\underset{\bm{\alpha}_{(q)}}{\text{min}}||\bm{\alpha}_{(q)}||_0, \qquad \text{such that }\mathbf{s}_{(q)} = \Phi_{(p)} \bm{\alpha}_{(q)}
		\end{equation*}
		are not sparse (i.e., $||\bm{\alpha}_{(q)}||_0$ is large). 
	\end{enumerate}
	With these assumptions and dictionaries, MCA solves the following optimization problem: 
	\begin{equation}\label{eq:MCA}
		\begin{split}
			\{\bm{\alpha}_{(1)}^{opt}, \hdots , \bm{\alpha}_{(k)}^{opt}\} &= \text{arg }\underset{\{\bm{\alpha}_{(1)}, \hdots , \bm{\alpha}_{(k)}\}}{\text{min}}\sum_{p = 1}^k ||\bm{\alpha}_{(p)}||_0  \\
			& \text{satisfying }\mbf{s} = \sum_{p = 1}^k \Phi_{(p)} \bm{\alpha}_{(p)}\\
		\end{split}
	\end{equation}
	MCA then determines the sources, $\{\mbf{s}_{(1)}, \hdots, \mbf{s}_{(k)}\}$, by projecting the estimated set of coefficients, $\{\bm{\alpha}_{(1)}^{opt}, \hdots , \bm{\alpha}_{(k)}^{opt}\}$, onto the dictionaries, $\{\Phi_{(1)}, \hdots, \Phi_{(k)}\}$.
	
	There are two options for finding the dictionaries to use with MCA. Firstly, sparsifying mathematical models that have quick forward and backward transforms can be used as dictionaries. Mathematical transforms often used are the Gabor transform, wavelets, short time Fourier transforms, and ridgelets \cite{MCA}. Whereas the Fourier dictionary sparsely represents smooth signals, the wavelet dictionary may sparsely represent piecewise smooth signals with point singularities providing a means of solving specific BSS problems with these morphologies present. This is the preferred method in MCA's development, though it requires human expertise to find the optimal transforms to use, and sometimes the morphologies present are intricate enough that mathematical transforms are ineffective. 
	
	Secondly, data-driven dictionaries are those that are estimated in an iterative fashion on a training dataset or the signal itself. Current state of the art examples include generalized PCA, K-SVD, and the union of Orthobases \cite{Dictionaries}. Due to their required training, the dictionaries learned from these data-driven techniques often generalize poorly. Additionally, there is often no accompanying procedure to separate the learned dictionary into multiple dictionaries for MCA applications. 
	
	Our method of using DMD to generate dictionaries for MCA avoids these issues because it requires no training on datasets beyond the signal to be decomposed. DMCA solely relies on the blind decomposition of the video's dynamics to generate multiple dictionaries representing different morphologies present in the individual frames. We offer a new method of creating dictionaries discovered fully from the video to be separated.

	\section{Dynamic morphological component analysis}\label{sec:dmca}
	
	To use DMD on our input video, we unroll each video frame into a column of a data matrix. In this format, the index of a column represents the time at which the frame was captured, and the columns are organized chronologically from left to right. In DMCA, we populate dictionaries with modes returned by doing DMD on skinny submatrices of our data matrix. As in MCA, each dictionary of ours must be designed so that its atoms efficiently represent a single morphology in the signal to be reconstructed. \textit{We define the morphology that each mode represents based on clustering in DMD eigenspace}. To separate a video into $k$ layers each with unique morphologies, we separate DMD eigenvalues into $k$ clusters on the complex plane, and assign modes to different dictionaries based on the clusters that their associated eigenvalues are assigned to. If two modes have eigenvalues clustered together on the complex plane, those two modes will be used as atoms in the same dictionary. 
	
	Clustering DMD modes by their eigenvalues is a natural way to distinguish between modes that represent different fundamental behavior. Given the DMD column reconstruction of $\mathbf{x}_t$ from \eqref{eq:dmdsd}, the coefficient representing how much a mode, $\bm{\varphi}_{j} := \Phi_{:, j}$, contributes in the decomposition of $\mathbf{x}_t$ is 
	\begin{equation}\label{eq:dynamics}
		a_j(t) = \Lambda_{j, j}^{t - 1}b_j,
	\end{equation} 
	which we call the dynamics of the mode $\bm{\varphi}_j$ at time $t$. The magnitude of this quantity communicates how dominant the structure $\bm{\varphi}_j$ is in the reconstruction of our measurements over time. Defining $\bm{\lambda} = \text{diag}(\Lambda)$ and with some algebraic manipulation \eqref{eq:dynamics} becomes 
	\begin{align}\label{eq:dynamics2}
		a_j(t) &= b_je^{(t-1) \ln|\lambda_j|}( \cos[(t-1)\arg(\lambda_j)]\\
		&+ i\sin[(t-1)\arg(\lambda_j)] ) \nonumber
	\end{align}
	where $\arg(\lambda_j)$ is the angle between the positive $x$-axis (representing the positive real numbers in the complex plane) and the line connecting $\lambda_j$ to the origin, and $i = \sqrt{-1}$. From this representation, it is apparent that eigenvalues closest to the imaginary axis will have a higher oscillatory frequency, modes with large magnitude eigenvalues will have exponentially increasing dynamics, and modes with low magnitude eigenvalues will have dynamics that decay rapidly. The position of $\lambda_j$ on the complex plane completely characterizes the behavior of $\bm{\varphi}_j$ in the reconstruction of the measurements. By clustering eigenvalues, we cluster modes with similar dynamics. 
	
	We now present our novel DMCA algorithm. Given a video unrolled into a data matrix, $\mathbf{X} \in \mathbb{C}^{m \times n}$, our problem is to separate that data matrix into $k$ distinct matrices, each of which is a video representing a different morphology present in the input video, and collectively sum to the original data matrix. Under the assumption that $\mathbf{X}$ is composed of at least one layer of texture and a target, we wish to separate $\mathbf{X}$ linearly as 
	\begin{equation}\label{eq:first_objective}
		\mathbf{X} = \mathbf{X}^{out}_{target} + \sum_{p \in \left\{\substack{\text{additional}\\ \text{layers}}\right\}} \mathbf{X}^{out}_{(p)}.
	\end{equation} 
	
	We offer a new method for linearly decomposing videos of this class into $k$ videos with distinct morphologies using the video's dynamics and scene-content variation. Our novel DMCA algorithm is outlined in the following steps:
	
	\begin{enumerate}
		\item Use a \textit{sliding DMD window} on the entire data matrix to generate a collection of modes and their corresponding eigenvalues. The modes will be the atoms in the dictionaries, and their eigenvalues will be used to separate the modes into their distinct dictionaries. This is outlined in Section \ref{sec:dmca_1}.
		\item Group eigenvalues based on the morphology they represent. Assign these labels to the corresponding modes. This is outlined in Section \ref{sec:dmca_2}.
		\item For each unrolled frame in the input video, define a subset of the modes that were taken from DMD windows near to or including that frame. Separate these modes into $k$ dictionaries by their labels. These will be the dictionaries used to do MCA on that video frame. This is outlined in Section \ref{sec:dmca_3}.
		\item Use a separation algorithm 
		to reconstruct each frame of the input video with elements from its $k$ different dictionaries. Once this is done for every frame of the input video, each frame is decomposed as the sum of atoms from dictionaries representing each distinct morphology, and the collections of these decomposed frames give us $k$ videos representing the distinct morphologies. This is outlined in Section \ref{sec:dmca_4}.
		 
	\end{enumerate}
	Steps 1 and 2 are the dictionary creation part of DMCA while steps 3 and 4 are the video reconstruction part. There is no communication between steps 1 \& 2 and steps 3 \& 4, and italicized terms will be defined later in this section. We will construct DMCA formally with each step having its own subsection. 
	
	\subsection{Sliding DMD window}\label{sec:dmca_1}
	
	DMD seeks to decompose a time-series matrix, which in our case we call $\mathbf{X} \in \mathbb{C}^{m \times n}$, into coherent structures that exist for all times that the data is recorded. However, in data matrices, we know that many coherent structures only exist in strict subspaces of the whole data matrix. That is, they do not necessarily exist in the video for all time. It is because of this that modes extracted from subsets of the data matrix, $\tilde{\mathbf{X}} \in \mathbb{C}^{m \times w} \text{ such that } (w < n)$, will be much different than those extracted from the whole data matrix, $\mathbf{X}$, and the frequency spectrum of modes from $\tilde{\mathbf{X}}$ will be different than those from DMD on all of $\mathbf{X}$. Doing DMD on a windowed subset of the data matrix allows us to tune the sample length to best capture the different timescales of the data.
	
	We define a sliding DMD window similar to that which is used in \cite{Dylewsky_2019} with some important modifications. A \textit{DMD window} is an $m \times w_L$ sub-matrix of $\mathbf{X}$ for some $w_L < n$. That is, if the columns of $\mathbf{X}$ are indexed as 
	\begin{equation*}
		\mathbf{X} = \begin{bmatrix}\mathbf{x}_1 & \mathbf{x}_2 & \hdots & \mathbf{x}_n \end{bmatrix},
	\end{equation*}
	then the $j^{th}$ DMD window, $\mathbf{X}^{win}_{(j)}$, is defined as 
	\begin{equation*}
		\mathbf{X}^{win}_{(j)} := \mathbf{X}_{:, [j, j + w_L - 1]} =  \begin{bmatrix}\mathbf{x}_j & \mathbf{x}_{j + 1} & \hdots & \mathbf{x}_{j + w_L - 1}\end{bmatrix}.
	\end{equation*}
	We use the notation of a matrix with a superscript in parentheses to designate DMD windows as to ensure that our notation for indexed submatrices is distinguishable from that of the DMD windows. 
	
	For each $j$ satisfying $1 \leq j \leq n - w_L + 1$, we do a full-rank DMD on the window $\mathbf{X}^{win}_{(j)}$ to extract modes and their associated eigenvalues. More specifically, using the notation from \eqref{eq:scripts} for each $\mathbf{X}^{win}_{(j)}$, we use DMD to find the spectral decomposition of the best-fit operator, $A_{(j)}$, which satisfies 
	\begin{align*} 	
		A_{(j)} = \underset{A}{\text{arg }\min}&||(\mathbf{X}^{win}_{(j)})_{:, [2, w_L]} - A(\mathbf{X}^{win}_{(j)})_{:, [1, w_L - 1]}||_F\\ 
		&\text{ where rank}(A) = w_L - 1
	\end{align*}
	This outputs modes, eigenvalues, and coefficients written as the triplets $\{(\bm{\varphi}_{(i, j)}, \lambda_{(i, j)}, b_{(i, j)})\}_{i = 1}^{w_L - 1}$, where the index $i, j$ indicates that the object belongs to the $i^{th}$ mode taken from the $j^{th}$ DMD window on $\mathbf{X}$. With this formulation, the triplets reconstruct the $k^{th}$ column of $\mathbf{X}^{win}_{(j)}$ as 
	\begin{equation}\label{eq:recon_window}
		(\mathbf{X}^{win}_{(j)})_{:, k} = \sum_{i = 1}^{w_L -1} \bm{\varphi}_{(i, j)} \lambda_{(i, j)}^{k - 1}b_{(i, j)}.
	\end{equation}
	We do DMD $n - w_L + 1$ different times to construct the collections 
	\begin{equation}
		\{\bm{\varphi}_{(i,j)}\}_{i \in \mathcal{I}_{i}, j \in \mathcal{I}_j}, \qquad \{\lambda_{(i, j)}\}_{i \in \mathcal{I}_{i}, j \in \mathcal{I}_j}
	\end{equation}
	where $\mathcal{I}_i = [1, w_L - 1]$ is the interval of indexes for which the DMD modes are defined, and $\mathcal{I}_j = [1, n - w_L + 1]$ is the interval of indexes for which the DMD windows are defined. 
	
	The collection $\{\bm{\varphi}_{(i,j)}\}$ contains all modes extracted from DMD windows across $\mathbf{X}$, which will be the atoms in our multiple dictionaries, and each element, $\lambda_{(i, j)}$, of $\{\lambda_{(i,j)}\}$ is the eigenvalue associated with the mode $\bm{\varphi}_{(i, j)}$, which is the $i^{th}$ mode taken from the DMD window $\mathbf{X}^{win}_{(j)}$.

	\subsection{Eigenvalue clustering}\label{sec:dmca_2}
	
	Next, we separate the elements of $\{\lambda_{(i, j)}\}$ into $k$ clusters. Once we have done this, we label each vector $\bm{\varphi}_{(i, j)}$ with the label assigned to $\lambda_{(i, j)}$. Three options for how to do so are included below, however, finding an optimal clustering method is a topic of future research.  
	
	\subsubsection{Magnitude threshold clustering}
	
	A simple method of clustering, as is done in \cite{grosek2014dynamicmodedecompositionrealtime}, is to separate modes solely by the magnitudes of their eigenvalues. The more ``stationary'' a mode is, the closer its associated eigenvalue is to the origin. Often, as will be shown in the examples section of this paper, clusters are obvious to the human eye, and, for the separation of eigenvalues into $k$ clusters, we can design a strictly decreasing sequence $s_{(1)} > s_{(2)} > \hdots > s_{(k)}$ that defines a labeling function $$\mathcal{L}(\lambda) = \max \, p \in \{1, 2, ..., k\} \,\,\, \text{such that  } s_p > |\lambda|^2$$to label each eigenvalue-mode pairing. This process requires human intervention, and in most cases, can be replaced with a k-medians or k-means clustering algorithm on the scalar magnitudes of the eigenvalues. However, in some cases, k-means and k-medians do not converge to properly segment the eigenvalues, whether due to $k$ misrepresenting the actual number of clusters or the existence of clusters that overlap. In these cases, and when using DMCA in a new scenario, manual clustering can be easy, explainable, and very effective.   
	
	\subsubsection{Radial clustering}
	
	Most eigenvalues are contained in the unit circle and exist in conjugate pairs. Because they dictate nearly morphologically identical dynamics, we would like the eigenvalues $\lambda_{(1)} = a + bi$ and $\lambda_{(2)} = a - bi$ to be clustered in the same group. Also, considering \eqref{eq:dynamics2}, the angle that an eigenvalue makes with the positive real axis in the complex plane contains much information about its morphology. We manually create a labeling function that uses this information. For $k$ bins, we define the labeling function $\mathcal{L}: \mathbb{C} \rightarrow \{1, \hdots , k\}$ as
	\begin{equation}\label{eq:easy_eigs}
		\mathcal{L}(a + bi) =  \lceil (k + 1) \arg(a + |b|i)/ \pi \rceil 
	\end{equation}
	where $(\lceil \cdot \rceil)$ is the ceiling function. Then, our eigenvalues that represent target modes are labeled in group 1, and modes with varying mixtures of growth, decay, and oscillation have labels between 1 and $k$.  
	
	\subsubsection{K-means or K-medians}
	
	As mentioned above, we feed the magnitudes of eigenvalues into a K-means or K-medians clustering algorithm. K-medians is used instead of K-means to cluster DMD eigenvalues in \cite{Dylewsky_2019} as not to inflate the separation of higher frequencies. This approach is more robust than the manual methods above because it will never return an empty cluster, which would result in a blank frame being part of the eventual reconstruction. However, in practice, often the cluster centers returned by K-medians and K-means are too near to another, and manual clustering is a more precise and explainable option.\\

	Once clustering is complete, we can apply a smoothing filter to the target modes (those with associated eigenvalues labeled with cluster 1) to remove residual noise in the final target video. This is optional, but in practice, using the Savitzky-Golay filter on the target modes has enhanced the signal to clutter ratio of our resulting target video.

	\subsection{Forming dictionaries}\label{sec:dmca_3}
	For each column, $\mathbf{x}_j$, of the inputted data matrix, we create a set of $k$  dictionaries, $\{\Phi_{(p, j)}\}_{p \leq k}$, to decompose $\mathbf{x}_j$ as a linear combination of elements from each dictionary as 
	\begin{equation}\label{eq:minimizethis}
		\mathbf{x}_j = \sum_{p = 1}^k \Phi_{(p, j)} \bm{\alpha}^{opt}_{(p, j)}
	\end{equation}
	with some coefficients $\{\bm{\alpha}_{(p, j)}\}_{p \leq k}$. We do not simply generate $k$ generic dictionaries (which would be the $k$ labeled matrices formed from the elements of $\{\bm{\varphi}_{(i, j)}\}$) to use for decomposing every column of $\mathbf{X}$. These dictionaries would be too large to perform quick computations and would contain many atoms that would go unused in reconstructing the column. Because coherent structures in a data matrix are often transient, we do not expect modes extracted from one window of the data matrix to effectively reconstruct a column on another side of that data matrix. So, it is advantageous to reconstruct each column of the data matrix with modes taken from nearby DMD windows. We construct dictionaries, $\{\Phi_{(p,j)}\}_{p \leq k}$, for each column $\mathbf{x}_j$ as follows. 
	
	We take the integer $w_N \in \mathbb{Z}_+$ as an input to DMCA, which tells us for a given column at a given index how many nearby DMD windows we want to use to construct the local dictionaries. For some $\mathbf{x}_j$, if $w_N = 10$, then we use modes from every window $\mathbf{X}^{win}_{(q)}$ satisfying $j - 10 \leq q \leq j + 10$ as atoms in our $k$ dictionaries. Using the labeling function, $\mathcal{L}: \mathbb{C} \rightarrow \{ 1, 2, \hdots, k\}$, we construct each dictionary, $\Phi_{(p, j)}$, as the matrix whose columns are composed of 
	\begin{equation}
		\text{every }\bm{\varphi}_{(i, q)} \text{ such that $|j - q| \leq w_N$ and $\mathcal{L}(\lambda_{(i,q)}) = p$}.
	\end{equation} 
	For every $\mathbf{x}_j$, because we specify that the linear operator, $A_{(j)}$, which linearly transforms $(\mathbf{X}^{win}_{(j)})_{:, [1, w_L - 1]}$ to $(\mathbf{X}^{win}_{(j)})_{:, [2, w_L]}$, is of full rank, $A_{(j)}$ is exactly
	$$A_{(j)} = (\mathbf{X}^{win}_{(j)})_{:, [2, w_L]} (\mathbf{X}^{win}_{(j)})_{:, [1, w_L - 1]}^{\dagger},$$and consequently,
	\begin{equation}\label{eq:inspan}
		\text{col}((\mathbf{X}^{win}_{(j)})_{:, [1, w_L - 1]}) = \text{col}(A_{(j)}) = \text{span}(\{\bm{\varphi}_{(i, j)}\}_{i \leq w_L -1}).
	\end{equation}
	Because each $\mathbf{x}_j$ is in multiple windows whose modes are the elements of $\{\Phi_{(p, j)}\}$, it follows from \eqref{eq:inspan} that for all $j \leq n$,
	\begin{equation}
		\mathbf{x}_j \in \bigcup_{p = 1}^k \text{col}(\Phi_{(p, j)}).
	\end{equation}
	Although each vector space, $\text{col}(\Phi_{(p, j)})$, may not be overcomplete for $\mathbb{C}^m$, we are always guaranteed that there exists some $\{\bm{\alpha}^{opt}_{(p, j)}\}_{p \leq k}$ satisfying $$\left|\mathbf{x}_j - \sum_{p = 1}^k \Phi_{(p, j)} \bm{\alpha}^{opt}_{(p, j)}\right| = 0.$$

	\subsection{Video reconstruction}\label{sec:dmca_4}
	Finally, we must use our created dictionaries to decompose $\mathbf{X}$ into $k$ different layers. For the sake of algorithmic implementation, for each signal $\mathbf{x}_j$, we reformulate our $k$ dictionaries into a large dictionary (for which we know the label of each atom). We define the dictionary $\Phi_{(j)}$ as the matrix whose columns are composed of $$\text{every } \bm{\varphi}_{(i, q)} \text{ where }|j - q| \leq w_N,$$ which is the matrix containing all modes extracted from DMD windows less than $w_N$ indexes away from $\mathbf{x}_j$. We notice that as long as $w_N \geq w_L$, which in practice is often the case (though need not be), there are at least $w_L$ distinct submatrices of $\Phi_{(j)}$ whose columns can be linearly combined to reconstruct $\mathbf{x}_j$ (except for the columns such that $|j - n| >  w_L$ or $j < w_L$). This follows from the fact that there are $w_L$ distinct sets of modes extracted from $w_L$ different full-rank DMD processes done on windows of $\mathbf{X}$ that contain $\mathbf{x}_j$. Consequently, the least squares problem
	\begin{equation}\label{eq:leastsquares}
		\underset{\bm{\alpha}_{(j)}}{\text{arg min}}||\mathbf{x}_j - \Phi_{(j)} \bm{\alpha}_{(j)}||_2 
	\end{equation}
	will have multiple solutions, $\bm{\alpha}^{opt}_{(j)}$, that diverge to infinity. Note that \eqref{eq:leastsquares} is equivalent to minimizing the $\ell_2$ norm of the error for the reconstruction of $\mathbf{x}_j$ in \eqref{eq:minimizethis}. Given that we are operating on frames with finite pixel values, we want our algorithm to converge to solutions that are finite. So, we include a penalizing term in our objective function to avoid solutions that diverge to infinity.
	
	For each column, $\mathbf{x}_j$, with the definitions of $\Phi_{(j)}$ and $\Phi_{(p, j)}$ from above, we pose the optimization problem
	\begin{equation}\label{eq:simpleopt}
		\begin{split}
			\{\bm{\alpha}^{opt}_{(p,j)}\} = \underset{\{\bm{\alpha}_{(p,j)}\}}{\text{arg min}} & \sum_{p = 1}^{k}   \left|\left|\mathbf{x}_j - \sum_{p = 1}^{k}\Phi_{(p,j)} \bm{\alpha}_{(p,j)}\right|\right|_2 \\ 
			&+\sum_{p = 1}^{k} \gamma||\bm{\alpha}_{(p, j)}||_1 
		\end{split}
	\end{equation}
	for some weight $\gamma > 0$. This problem is equivalently formulated as 
	\begin{equation}\label{eq:optimize}
		\bm{\alpha}^{opt}_{(j)} = \underset{\bm{\alpha}_{(j)}}{\text{arg min}} \left|\left|\mathbf{x}_j - \Phi_{(j)} \bm{\alpha}_{(j)}\right|\right|_2 +  \gamma ||\bm{\alpha}_{(j)}||_1.
	\end{equation}
	The solution to this problem remains bounded, and there are several algorithms that find it efficiently.  
	
	Upon the estimation of $\bm{\alpha}^{opt}_{(p, j)}$ for every $1 \leq j \leq n$ and $1 \leq p \leq k$, we reconstruct each video $\mathbf{X}^{out}_p$ as 
	\begin{equation}\label{eq:final_data matrix}
		\mathbf{X}^{out}_p = \begin{bmatrix}
			\Phi_{(p,1)} \bm{\alpha}^{opt}_{(p, 1)} & \Phi_{(i, 2)} \bm{\alpha}^{opt}_{(p, 2)} & \hdots & \Phi_{(p, n)} \bm{\alpha}^{opt}_{(p, n)}
		\end{bmatrix}.
	\end{equation}
	
	Each data matrix, $\mathbf{X}^{opt}_p$, whose columns are flattened video frames, is composed entirely of DMD modes which are grouped together in DMD eigenspace and assigned the label ``$p$''. Additionally, each column of $\mathbf{X}^{opt}_p$ is the sum of modes that are captured from DMD windows on the input video that are temporally close. With this localized reconstruction, each of our output videos is defined by the modes whose eigenvalues fall in a certain cluster in DMD eigenspace and are captured from nearby frames. In many instances, this process of mode discrimination captures morphologies present in the input video. All together, the DMCA algorithm is stated as Algorithm 1.

	\begin{algorithm}[H]
		\caption{Dynamic Morphological Component Analysis}\label{alg:DMCA}
		\begin{algorithmic}
			\STATE \textbf{Input: } $\mathbf{X} \in \mathbb{C}^{m \times n}$ (data matrix), $w_L \in \mathbb{Z}_+$ (DMD window length), $\mathcal{L}: \mathbb{C} \rightarrow \mathbb{Z}$ (labeling function), $w_N \in \mathbb{Z}_+$ (reconstruction parameter).
			\STATE \textbf{Step 1: Dictionary Generation}
			\FOR {$j \in \{1,2,\ldots, n - w_L + 1\}$}
			\STATE $\mathbf{X}_{\text{window}} \leftarrow \left[\mathbf{x}_{j} \,\, \hdots \,\, \mathbf{x}_{j + w_L - 1} \right]$.
			\vspace{0.4em}
			\STATE Do DMD on $\mathbf{X}_{\text{window}}$ to generate DMD modes, $\{\bm{\varphi}_{(i, j)}\}_{i = 1}^{w_L - 1}$, and DMD eigenvalues,  $\{\lambda_{(i, j)}\}_{i = 1}^{w_L -1 }$.
			\vspace{0.4em}
			\ENDFOR
			\STATE \textbf{Step 2: Component Separation}
			\STATE $k \leftarrow |\{\mathcal{L}(\lambda)\, | \, \lambda \in \mathbf{\lambda}\}|$ ($\#$ of output labels/layers)
			\FOR {$j \in \{1,2,\ldots, n\}$}
			\STATE Solve \eqref{eq:simpleopt} for the coefficients $\{\bm{\alpha}^{opt}_{(p, j)}\}_{p = 1}^k$ \\
			\FOR {$p \in \{1, \hdots, k\}$} 
			\STATE Append the output data matrix, $\mathbf{X}^{out}_p$, with the column-vector $\Phi_{(p, j)}\bm{\alpha}^{opt}_{(p, j)}$. 
			\ENDFOR
			\ENDFOR

			\STATE \hspace{0.5cm}\textbf{return} resultant component matrices, $\{\mathbf{X}^{out}_1, ..., \mathbf{X}^{out}_k\}$
		\end{algorithmic}
	\end{algorithm}

	\subsection{Motivating Example}
	
	While DMCA is designed to decompose videos arranged into data-matrices, a motivating example for how the DMD process works can be done on a still image. A still image that is the sum of multiple layers can be input to Alg. \ref{alg:DMCA} wherein it will be treated as a video whose columns are the frames flattened into one-dimensional vectors. For most images, this treatment as a dynamical process is not natural, and DMD offers little use separating the morphologies present. However, when we consider an image summed with a grid, the grid's morphology is the same as a video that oscillates from frame to frame, and thus can be separated from the image using DMCA. 
	
	In this example of a still image, the sliding DMD window will generate modes by sliding from left to right across the image, then it will sort these modes into two dictionaries based on their eigenvalues, and finally it will use these modes as atoms to reconstruct each column in the image.    
	
	\begin{figure}[h]
		\centering
		\includegraphics[width= 3 in]{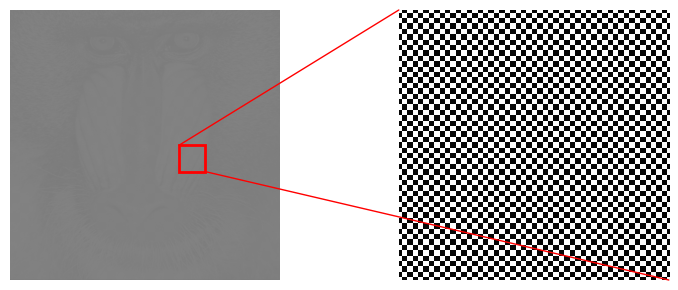}
		\caption{Still image of a mandrill summed with a grid inputted to DMCA.}
		\label{fig:input_still}
	\end{figure}
	
	As an experiment, we sum an image of a mandrill, in which the pixels range in intensity from 0 to 255, with a checkerboard image where the pixels alternate between having intensities of -3500 and 3500. Then, we scale the resultant photo so that its pixels range between 0 and 255 again. This input photo, as shown in Fig. \ref{fig:input_still}, has a PSNR of 5.69 dB and a structural similarity index measure (SSIM) of 0.0061. 
	
	We apply DMCA with a window length of 12, the labeling function 
	\begin{equation}
		\mathcal{L}(\lambda) = \begin{cases}
			1 & \qquad \text{if }|\lambda + 1|^2 > 0.01 \\
			2   & \qquad \text{if }|\lambda + 1|^2 \leq 0.01,
		\end{cases}
	\end{equation}
	and the parameter $w_N = 20$ to reconstruct two images: one representing the target and one representing the texture.
	\begin{figure}[h]
		\centering
		\includegraphics[width= 1.5 in]{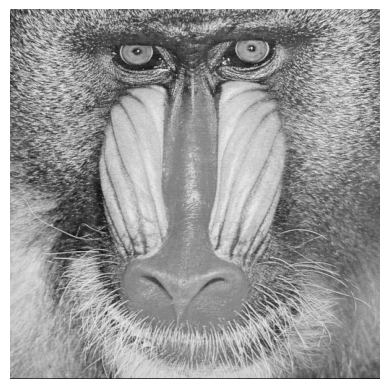}
		\caption{DMCA reconstructed image of the mandrill.}
		\label{fig:Lena_out}
	\end{figure}
	\begin{figure}[h]
		\centering
		\includegraphics[width= 2.7 in]{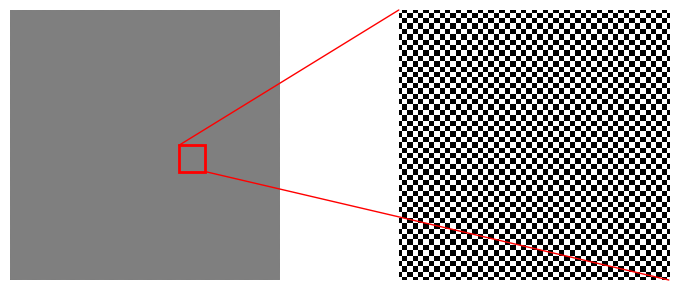}
		\caption{DMCA reconstructed texture layer.}
		\label{fig:Blocks_out}
	\end{figure}
	
	The target image (from cluster 1) has a PSNR of 31.7 dB, an SSIM of 0.96, and is plotted in Fig. \ref{fig:Lena_out}, while the texture layer returned is plotted in Fig. \ref{fig:Blocks_out}. Thus, by learning and identifying the morphology of the checkerboard, DMCA was able to extract it from the input image, and return a high-fidelity image of the other morphology present, which was the mandrill. We apply this algorithm to video data in section \ref{sec:exp}.

	\section{Experimental Results}\label{sec:exp}

	To demonstrate DMCA's capabilities, we test it on a series of datasets and compare its results with those of baseline methods. First, we test DMCA on videos with additive noise sampled from non-Gaussian distributions whose parameters change over time. Then, we use DMCA to separate an ``X'' tracing a random walk from a video of a sea state's height from the model in \cite{RIZAEV2022120}. Finally, we apply DMCA to a sequence of ISAR images with wind-clutter. 
	
	\subsection{Video Denoising}
	
	Unlike many denoising algorithms that are developed to extract noise sampled from a specific distribution, DMCA is designed to separate layers from videos with distinct dynamics, which, in the context of noise removal, includes a very wide range of noise distributions. We demonstrate these broad capabilities by removing noise that is sampled from distributions that evolve in time according to the following scheme. 
	
	Each pixel of the noise layer, $I_{i, j}$, at a given frame $t$ is sampled from the sum of distributions 
	\begin{equation}
		I_{i, j}(t) = \sum_{m = 1}^6 X_m(t)
	\end{equation}
	where $X_1(t)$ and $X_2(t)$ are samples from normal distributions
	\begin{align*}
		X_1(t) & \sim \mathcal{N}(300, 100 \rho(\sin(2\pi t/6) + 1)\\
		X_2(t) & \sim \mathcal{N}(-400, 225 \rho (\sin(2\pi t/6 + \pi/3) + 1 ),
	\end{align*}
	$X_3(t)$ and $X_4(t)$ are samples from uniform distributions
	\begin{align*}
		X_3(t) & \sim  \mathcal{U}(-100 + 175 \rho (\sin(2\pi t/6 + 4\pi/3) + 1),\\
		& 300 - 175 \rho(\sin(2\pi t/6 + 4\pi/3) + 1) \nonumber \\
		X_4(t) & \sim \mathcal{U}(400 - 170 \rho(\sin(2\pi t/6 + 4\pi/3) + 1),\\
		& 400 + 170 \rho (\sin(2\pi t/6 + 4\pi/3) + 1)  ), \nonumber
	\end{align*}
	and $X_5(t)$ and $X_6(t)$ are samples from Laplacian distributions
	\begin{align*}
		X_5(t) & \sim \text{Laplace}(-250,  225 \rho(\sin(2\pi t/6 + 2\pi/3) + 1))\\
		X_6(t) & \sim \text{Laplace}(-350, 125 \rho (\sin(2\pi t/6 + 5\pi/3) + 1)).
	\end{align*}
	In the following examples, $\rho \in [0, 1]$ is a parameter to scale the intensity of the noise. The noise video is added to a video from the Adobe 240fps dataset \cite{su2024adobe240fps} whose pixels range from 0 to 255. Once the two videos are summed, the entire video is scaled so that the pixels again range from 0 to 255. 
	
	This scheme adds noise from a rapidly evolving non-Gaussian distribution. We compare DMCA's performance to Video Block-Matching 4D Flitering (V-BM4D) as a baseline, though V-BM4D is designed for additive Gaussian noise \cite{Maggioni2012}. 
	
	\begin{figure}[h]
		\centering
		\includegraphics[width=2.5in]{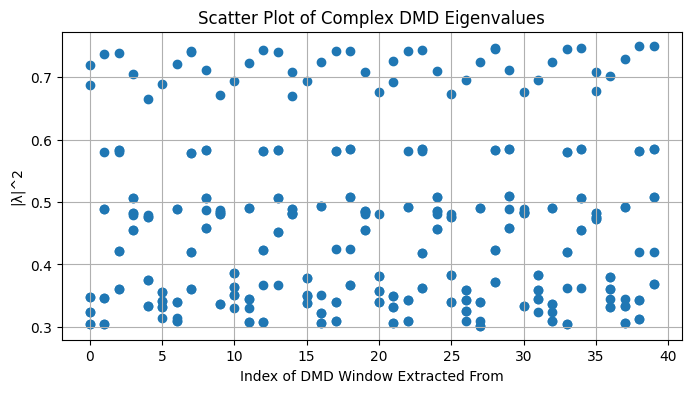}
		\caption{Plot of DMD eigenvalues magnitude squared in relation to the position of the DMD window they are extracted from.}
		\label{fig:eigenvalueplot}
	\end{figure}
	For $\rho = 0.5$ noise added to the video of the shipyard, and a window length of 8, DMCA returns the plot of DMD eigenvalues in Fig \ref{fig:eigenvalueplot}. From Fig. \ref{fig:eigenvalueplot}, it is apparent that the band of eigenvalues furthest from the x-axis represents the target, which in this case is the denoised video. We use the clustering function 
	\begin{equation}
		\mathcal{L}(\lambda) = \begin{cases}
			1 & \qquad \text{if }|\lambda|^2 > 0.6 \\
			2   & \qquad \text{if }|\lambda|^2 \leq 0.6
		\end{cases}
	\end{equation}
	with the parameter $w_N = 6$ on three videos from the Adobe dataset.

	\begin{figure*}[]
		\centering

		\begin{tabular}{>{\centering\arraybackslash}m{2cm} c c c}
			& (a) & (b) & (c) \\

			Base truth &
			\includegraphics[width=0.27\textwidth]{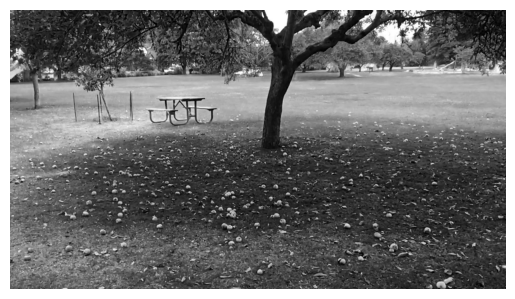} &
			\includegraphics[width=0.27\textwidth]{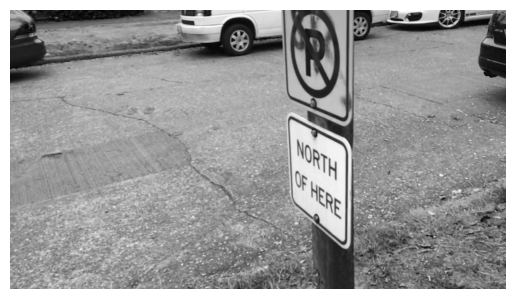} &
			\includegraphics[width=0.27\textwidth]{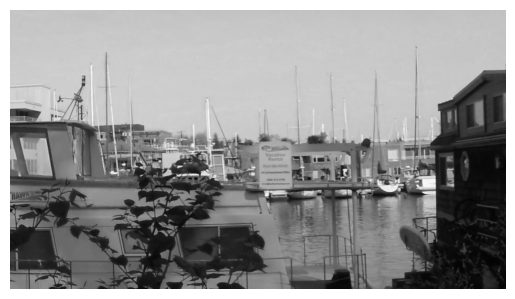} \\

			\shortstack{With noise\\$\rho$ = 0.5} &
			\includegraphics[width=0.27\textwidth]{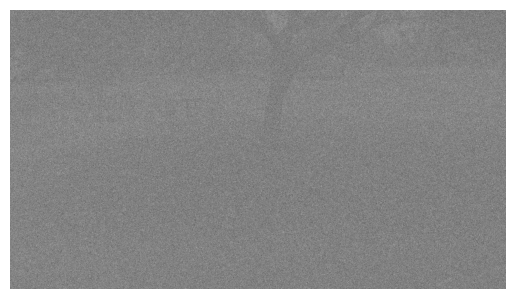} &
			\includegraphics[width=0.27\textwidth]{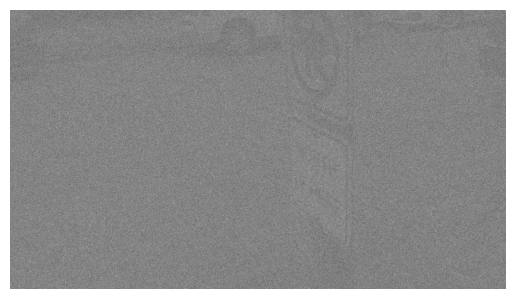} &
			\includegraphics[width=0.27\textwidth]{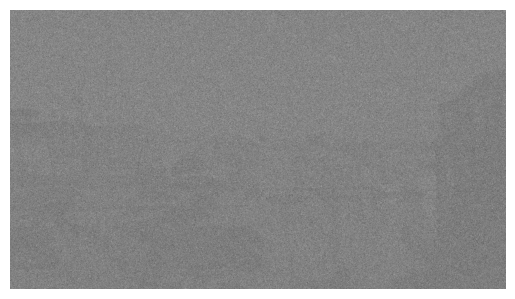} \\

			\shortstack{V-BM4D \\ output } &
			\includegraphics[width=0.27\textwidth]{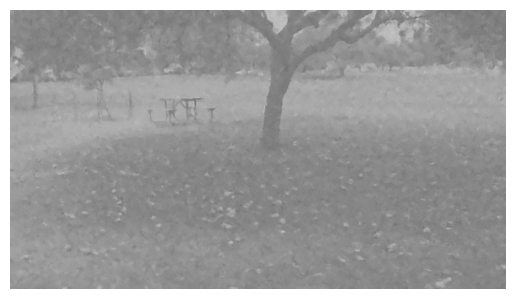} &
			\includegraphics[width=0.27\textwidth]{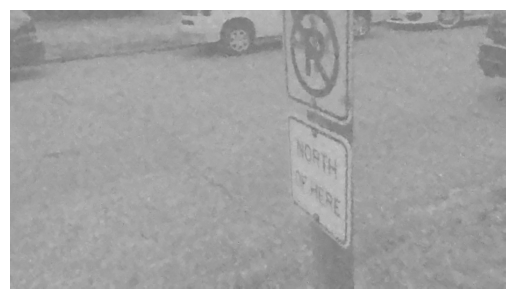} &
			\includegraphics[width=0.27\textwidth]{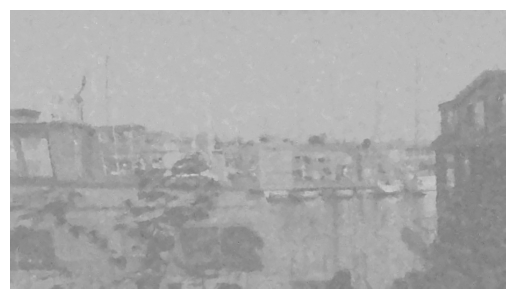} \\

			\shortstack{DMCA \\ output} &
			\includegraphics[width=0.27\textwidth]{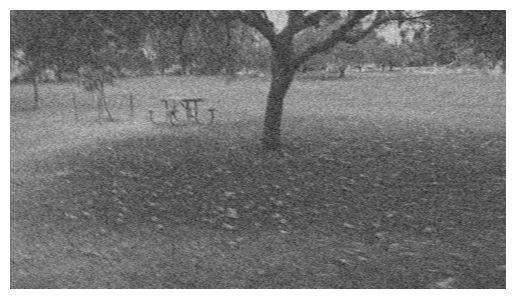} &
			\includegraphics[width=0.27\textwidth]{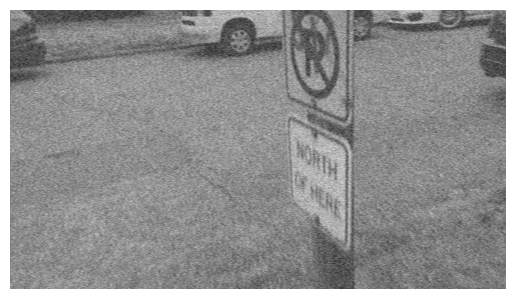} &
			\includegraphics[width=0.27\textwidth]{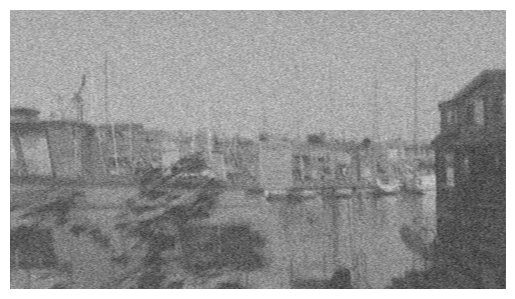} \\
		\end{tabular}
		
		\caption{Noise removal results from Adobe 240fps dataset.}
		\label{fig:denoising}
	\end{figure*}

	We compare the $30^{th}$ frame that DMCA and V-BM4D return when scaled back to ranging from 0 to 255 for the noisy video inputs in Fig. \ref{fig:denoising}. 
	\begin{figure}[h!]
		\centering
		\includegraphics[width=2.5in]{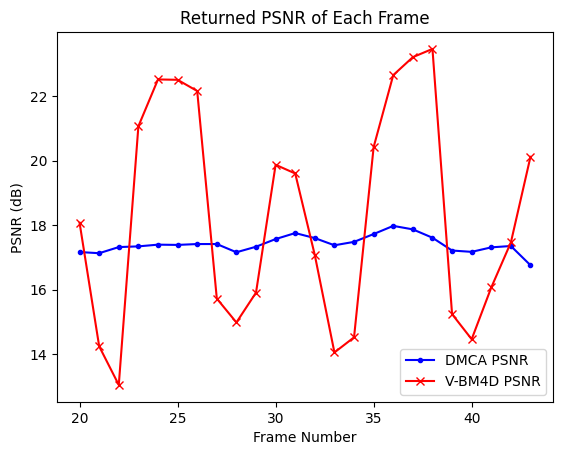}
		\caption{PSNR returned for shipyard video with noise intensity defined by $\rho = 0.5$.}
		\label{fig:PSNR_ship}
	\end{figure}
	For these particular frames, DMCA exhibits a higher PSNR than V-BM4D for each video, as seen in Fig. \ref{fig:PSNR_ship}, though they still have some noise artifacts in the target frames.
	
	Though V-BM4D is superior for many other frames, it proves much more volatile in noise removal throughout the video. In Fig. \ref{fig:PSNR_ship}, we plot the PSNR for 25 frames returned by DMCA and V-BM4D demonstrating this volatility. While V-BM4D offers a much higher maximum PSNR than DMCA throughout the frames, it also has a much lower minimum PSNR. While the quality of the video returned by V-BM4D fluctuates greatly depending on the distribution of noise at a given frame, DMCA's returned video does not.

	\begin{figure}[h]
		\centering
		\includegraphics[width=2.5in]{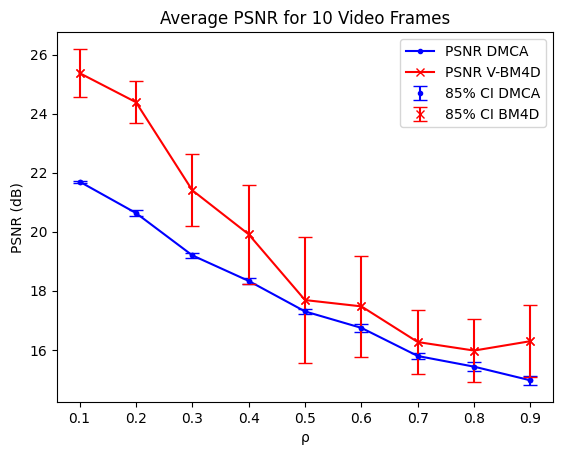}
		\caption{Average PSNR for Shipyard Videos with 85\% Confidence Interval.}
		\label{fig:avg_PSNR}
	\end{figure}
	
	This behavior is more evident when we plot the average PSNR and the 85\% confidence intervals of the PSNR for ten consecutive frames returned by V-BM4D and DMCA. While the frames returned by DMCA consistently have a lower average PSNR than those returned by V-BM4D, the confidence interval of DMCA is more compact for all $\rho$, demonstrating greater stability than the competitor.

	\subsection{Sea State Separation}
	
	Next, we provide an example of DMCA separating a moving target from the height map of an active sea state. In this experiment, we generate a height map of a sea state using the simulator from \cite{RIZAEV2022120} with a facet size of one meter, a scene size of 1000 meters, 10 m/s wind at 10 meters above sea-level, 50 km fetch, and a 35 degree angle of wind. We process this simulation in a video capturing a frame every $1/4$ second and cropping each frame to be 480x720 pixels. We then add a layer of Gaussian noise with a standard deviation of 0.02 meters and a video of our moving target, which is an `X' 20 pixels in both width and length with a height of 0.8 meters. 
	
	\begin{figure}[h]
		\centering
		\includegraphics[width= 3 in]{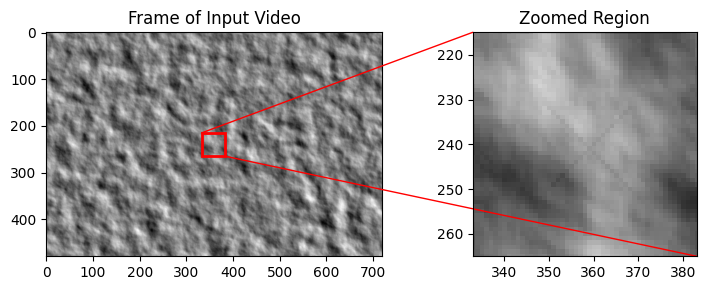}
		\caption{$40^{th}$ frame of video inputted to DMCA.}
		\label{fig:input_zoomed}
	\end{figure}
	
	The `X' is translated for every time increment according to a symmetric random walk where its center point, $(x_t, y_t)$, is moved to $(x_t + 1, y_t)$,  $(x_t - 1, y_t)$,  $(x_t, y_t + 1)$, $(x_t, y_t - 1)$, or remains unchanged, each with a probability of $1/5$. Once the three videos are summed, the resultant video is scaled so that pixels range from 0 to 255 and then inputted to DMCA. An image of an example frame of this video, and a zoomed in window displaying the `X,' is pictured in Fig. \ref{fig:input_zoomed}.  
	
	\begin{figure}[h]
		\centering
		\includegraphics[width= 3 in]{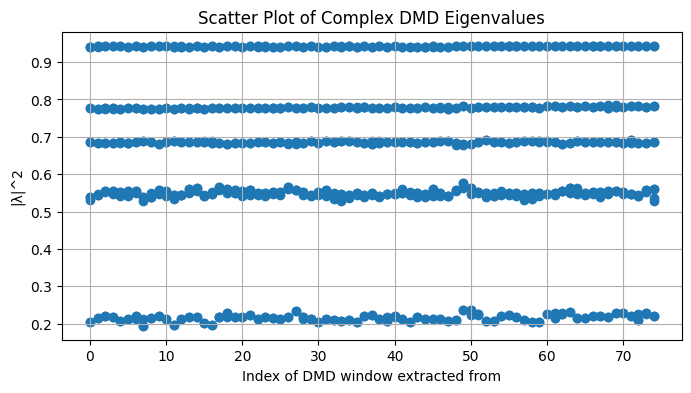}
		\caption{Plot of DMD eigenvalues magnitude squared in relation to the position of the DMD window they are extracted from.}
		\label{fig:2_eigenvalues}
	\end{figure}
	
	As seen from the figure, the target is hardly visible when summed with the sea state (SNR of -1.08 dB). We perform DMCA with a window length of 12, which returns the eigenvalue plot in Fig. \ref{fig:2_eigenvalues}. We then employ a k-medians clustering function on the collection of $|\lambda|^2$ with $k = 5$ clusters and the DMCA parameter $w_N = 2$. 
	
	\begin{figure*}[] 
		\centering
		
		\includegraphics[width=0.32\linewidth]{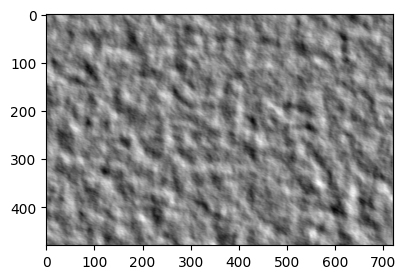} \hfill
		\includegraphics[width=0.32\linewidth]{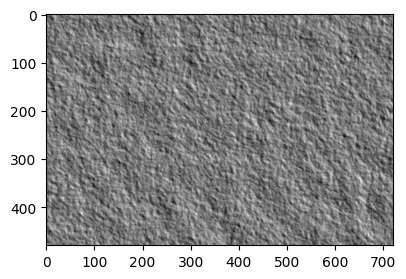} \hfill
		\includegraphics[width=0.32\linewidth]{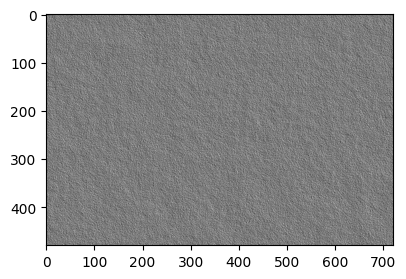}
		
		\vspace{0.2cm}
		
		\includegraphics[width=0.55\linewidth]{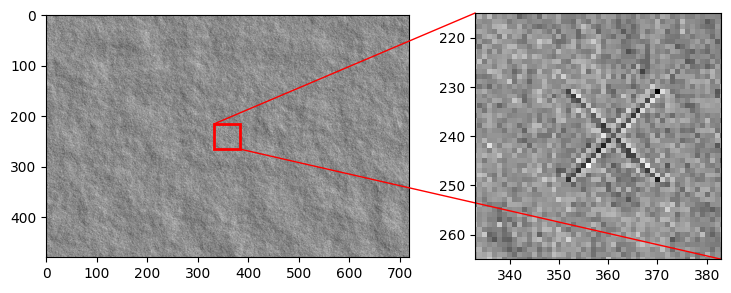} \hfill
		\includegraphics[width=0.32\linewidth]{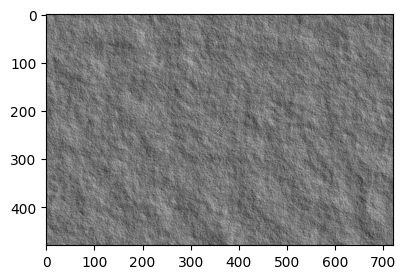}
		
		\caption{Top left: frame from DMCA returned video of cluster 1; top middle: frame from DMCA returned video of cluster 2; top right: frame from DMCA returned video of cluster 3; bottom left: frame from DMCA returned video of cluster 4; bottom right: frame from DMCA returned video of cluster 5.}
		\label{fig:5_returned}
		
	\end{figure*}
	
	This process returns five videos whose respective $40^{th}$ frames are displayed in Fig. \ref{fig:5_returned}. From this figure, it is seen that each cluster in DMD eigenspace is representative of a different component of the sea state and the target. Cluster 1 contains modes representing waves with larger wavelengths, likely corresponding to the gravity waves in the sea state model, while cluster 2 corresponds with modes representing the smaller wavelengths in the sea state model. Cluster 3 appears to be noise texture with some residue of waves, and cluster 4 contains very low magnitude waves and the target with an increased SNR of 5.86 dB. Cluster 5 represents larger low magnitude waves. 
	
	In this example, DMCA was able to enhance the SNR of a target in the presence of sea waves without any prior knowledge about the sea state or the target. Neither the spatial structure of the target nor the spatial structure of the sea state is leveraged for separation by DMCA; only the dynamics of the components are exploited for decomposition. Furthermore, no prior information about the dynamics of the components was used in the separation. The DMCA process discovered dictionaries for separation completely from the input video even when the clutter contained much more structure than additive noise. \\
	
	DMCA is equipped to handle a wide variety of modalities beyond real-valued denoising and sea state separation applications. We demonstrate the range of DMCA's separation abilities on synthetic, complex-valued ISAR data.

	\subsection{Inverse Synthetic Aperture Radar}
	
	Whereas synthetic aperture radar (SAR) uses the known motion of the sensor's platform to synthesize a large aperture for imaging, ISAR analyzes the Doppler frequency shifts of reflections from a target, which follows some complex trajectory, to model the target's rotational and translational motion. Along with additional processing and a stationary sensor platform, this analysis amounts to a synthetically enlarged aperture and an enhanced image resolution \cite{ISAR_book}.

	\begin{table}[]
		\caption{Radar Parameters for Generating ISAR Images \label{tab:table1}}
		\centering
		\begin{tabular}{||c | c||}
			\hline
			Parameters & Values  \\ [0.4ex] 
			\hline\hline
			Carrier frequency  & 77GHz\\
			\hline
			Stretch bandwidth  & 8MHz\\
			\hline
			Sampling Frequency  & 5MHz\\
			\hline
			Chirp Rate  & $60 \times 10^{12} \text{ Hz}^2$\\
			\hline
			Chirp Duration & 83.33 $\mu$s\\
			\hline
			Coherent Processing Interval & 0.1 s\\
			\hline
			Doppler Resolution & 10 Hz\\
			\hline
			Range Resolution & 0.075 m\\
			\hline
			Minimum Cross-Range Resolution & 0.19 m\\
			\hline
			Transmitted Power & 25dBm\\
			\hline
		\end{tabular}
	\end{table}
	
	We use simulated ISAR images of a moving bicycle with wind-clutter from the dataset \cite{hjw9-d428-21, 9695280}, whose parameters are recorded in Table \ref{tab:table1}. The dataset simulates ISAR images of identical targets with increasing wind speeds of (2.5, 5, 7.5, 10 m/s). For these sequences, signal-to-clutter ratio (SCR) increases with wind speed. Additionally, a range of ISAR image sequences representing an identical target are provided with additive Gaussian noise yielding sequences with (-5, 0, 5, 10dB) signal-to-noise ratios (SNR) and no wind-clutter.  
	
	We use the 10dB SNR ISAR image sequences to determine which pixels represent the target in each image. Since the trajectories of the targets of interest in the data with wind-clutter and with minimal additive noise (high SNR) are identical, we assume that the indexes of the cells containing the target in the high SNR image are identical to the cells containing the target in the images with wind-clutter, which we use in testing.  
	
	For each image in the ISAR sequences, we define the SCR as the ratio of the average target cell intensities to the average clutter cell intensities. Because we are estimating which cells contain the target from an image with noise, we cannot calculate the SCR with certainty. However, in experiments, the improvement in SCR from DMCA has proven to be independent of the threshold chosen for deciding which cells contain the target. 
	
	\begin{figure}[]
		\centering
		\includegraphics[width= 2.7 in]{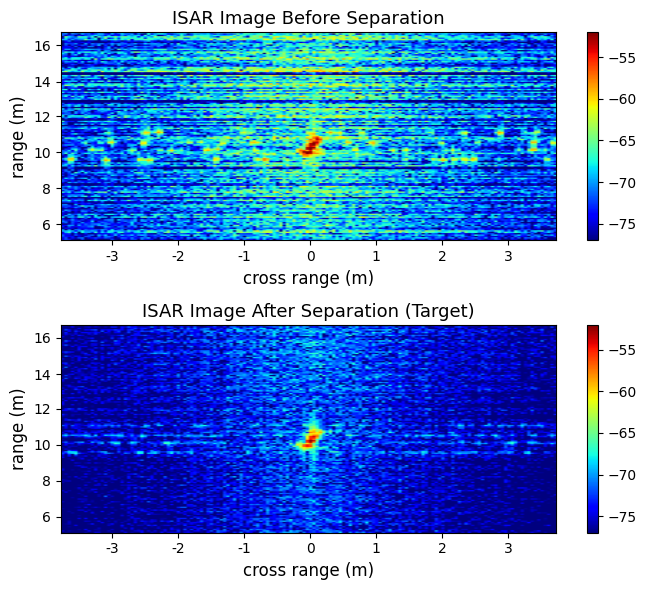}
		\caption{Separation of an ISAR frame from the sequence in \cite{hjw9-d428-21}.}
		\label{fig:ISAR_fig1}
	\end{figure}
	
	We demonstrate DMCA's effectiveness on the complex-valued ISAR sequence of a bicycle turning left towards the North with 10 m/s wind speed. On the complex-valued data matrix whose columns are the flattened ISAR images from the given sequence, we use a window length of 5, the parameter $w_N = 8$, and the labeling function 
	\begin{equation}
		\mathcal{L}(\lambda) = \begin{cases}
			1 & \qquad \text{if }|\lambda|^2 > 0.24 \\
			2   & \qquad \text{if }|\lambda|^2 \leq 0.24
		\end{cases}
	\end{equation}
	to reconstruct two ISAR image sequences, one representing the target, and one representing the wind-clutter. The $8^{th}$ frame input into DMCA and the returned frame representing the target are plotted in Fig. \ref{fig:ISAR_fig1}. For these frames, the input SCR is 24.8 dB and the output SCR is 28.0 dB.  
	
	As an experiment, we test MCA with two predefined dictionaries according to \cite{NRL_2017}. The first dictionary is a short-time Fourier transform (STFT) with a large window size, designed to sparsely represent the component in range cells with a narrow Doppler bandwidth, and the second dictionary is an STFT with a small window size designed to sparsely capture the wide Doppler bandwidth component of the range bins. Assuming that the target has a narrow bandwidth and that the clutter has a wider bandwidth, dictionary one should capture the target in the ISAR image, and dictionary two should capture the wind clutter. 
	
	After exhaustive searching for STFT window lengths and regularization parameters, we found that the predefined STFT dictionaries were unable to separate wind clutter from target in the simulated ISAR data. MCA, with these dictionaries, returned two components that were identical to the input up to multiplication by a scalar. 
	
	\begin{figure}[]
		\centering
		\includegraphics[width= 2.7 in]{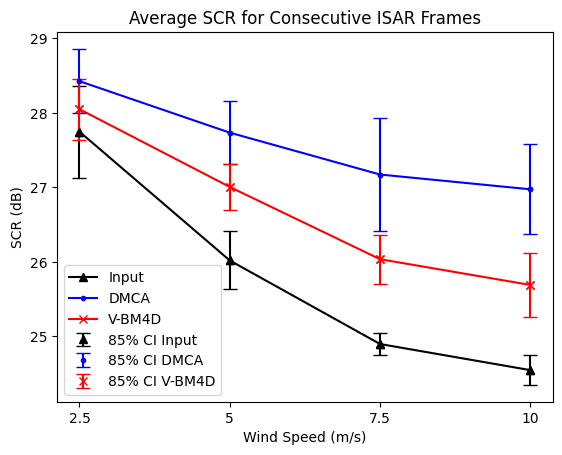}
		\caption{Target separation for six consecutive ISAR frames.}
		\label{fig:ISAR_fig2}
	\end{figure}
	
	To compare DMCA's clutter removal performance, we apply V-BM4D to the absolute value of the sequence of ISAR images in the synthetic dataset. We apply V-BM4D and DMCA to a sequence of six ISAR images for each wind speed, and record output SCR in Fig. \ref{fig:ISAR_fig2}. In this plot, the SCR is calculated from 30x30 pixel cropped images containing the target due to the small size of the target and a large scene. DMCA offers a consistent average increase in SCR from the input frames and the frames returned by BM4D for every wind speed, and has a confidence interval separated from that of the input ISAR image and that returned by V-BM4D for the two greatest wind speeds.

	\section{Conclusion}\label{sec:conclusion}
	
	When a video contains multiple layers that evolve incoherently over time, we empirically found that, when using sliding DMD windows, the DMD eigenvalues returned fall into distinct clusters even when the layers exhibit significant spatial overlapping. Just as MrDMD and its extension use clustering in DMD eigenspace to separate observations of physical systems based on their respective timescales, we use clustering in DMD eigenspace to separate our atoms into different dictionaries for the use of MCA. 
	
	In practice, this process has worked effectively when the layers have dissimilar dynamics, resulting in distinctly separated clusters in DMD eigenspace and multiple incoherent dictionaries. While we have yet to develop theory and related numerical bounds determining when DMCA solves BSS problems on videos effectively, we have empirically found that it is successful in many situations involving the separation of a smoothly evolving scene and noise, a target following a random walk and a sea state, and a target from wind-clutter in ISAR images. Additionally, DMCA has been shown to work when separating some periodic texture from a still image. 
	
	In future research, we hope to provide numerical bounds for DMCA's effectiveness and optimize the clustering algorithm for DMD eigenvalues.

	\section*{Acknowledgments}
	We would like to thank all reviewers for
	their thoughtful comments, which helped us improve this manuscript.
	
	\clearpage	
	
	\bibliographystyle{IEEEtran}
	\bibliography{refs.bib}

\begin{thebibliography}{10}
\providecommand{\url}[1]{#1}
\csname url@samestyle\endcsname
\providecommand{\newblock}{\relax}
\providecommand{\bibinfo}[2]{#2}
\providecommand{\BIBentrySTDinterwordspacing}{\spaceskip=0pt\relax}
\providecommand{\BIBentryALTinterwordstretchfactor}{4}
\providecommand{\BIBentryALTinterwordspacing}{\spaceskip=\fontdimen2\font plus
\BIBentryALTinterwordstretchfactor\fontdimen3\font minus
  \fontdimen4\font\relax}
\providecommand{\BIBforeignlanguage}[2]{{%
\expandafter\ifx\csname l@#1\endcsname\relax
\typeout{** WARNING: IEEEtran.bst: No hyphenation pattern has been}%
\typeout{** loaded for the language `#1'. Using the pattern for}%
\typeout{** the default language instead.}%
\else
\language=\csname l@#1\endcsname
\fi
#2}}
\providecommand{\BIBdecl}{\relax}
\BIBdecl

\bibitem{Kreyszig2007}
\BIBentryALTinterwordspacing
\emph{Introductory Functional Analysis with Applications}, ser. Wiley classics
  library.\hskip 1em plus 0.5em minus 0.4em\relax Wiley India Pvt. Limited,
  2007. [Online]. Available:
  \url{https://books.google.com/books?id=osXw-pRsptoC}
\BIBentrySTDinterwordspacing

\bibitem{oppenheim1975digital}
\BIBentryALTinterwordspacing
A.~Oppenheim and R.~Schafer, \emph{Digital Signal Processing}, ser. MIT video
  course.\hskip 1em plus 0.5em minus 0.4em\relax Prentice-Hall, 1975. [Online].
  Available: \url{https://books.google.com/books?id=vfdSAAAAMAAJ}
\BIBentrySTDinterwordspacing

\bibitem{mallat08}
S.~Mallat, \emph{A Wavelet Tour of Signal Processing, Third Edition: The Sparse
  Way}, 3rd~ed.\hskip 1em plus 0.5em minus 0.4em\relax USA: Academic Press,
  Inc., 2008.

\bibitem{Grochenig2001foundations}
\BIBentryALTinterwordspacing
K.~Gr{\"o}chenig, \emph{Foundations of Time-Frequency Analysis}, ser. Applied
  and Numerical Harmonic Analysis.\hskip 1em plus 0.5em minus 0.4em\relax
  Birkh{\"a}user Boston, 2001. [Online]. Available:
  \url{https://books.google.com/books?id=sjN2qq99-WwC}
\BIBentrySTDinterwordspacing

\bibitem{ksvd06}
M.~Aharon, M.~Elad, and A.~Bruckstein, ``K-svd: An algorithm for designing
  overcomplete dictionaries for sparse representation,'' \emph{IEEE
  Transactions on Signal Processing}, vol.~54, no.~11, pp. 4311--4322, 2006.

\bibitem{Dictionaries}
R.~Rubinstein, A.~M. Bruckstein, and M.~Elad, ``Dictionaries for sparse
  representation modeling,'' \emph{Proceedings of the IEEE}, vol.~98, no.~6,
  pp. 1045--1057, 2010.

\bibitem{pmlr-v9-jenatton10a}
R.~Jenatton, G.~Obozinski, and F.~Bach, ``Structured sparse principal component
  analysis,'' in \emph{Proceedings of the Thirteenth International Conference
  on Artificial Intelligence and Statistics}, ser. Proceedings of Machine
  Learning Research, Y.~W. Teh and M.~Titterington, Eds., vol.~9.\hskip 1em
  plus 0.5em minus 0.4em\relax Chia Laguna Resort, Sardinia, Italy: PMLR,
  13--15 May 2010, pp. 366--373.

\bibitem{sparse_dictionaries}
R.~Rubinstein, A.~M. Bruckstein, and M.~Elad, ``Dictionaries for sparse
  representation modeling,'' \emph{Proceedings of the IEEE}, vol.~98, no.~6,
  pp. 1045--1057, 2010.

\bibitem{wright09}
J.~Wright, A.~Y. Yang, A.~Ganesh, S.~S. Sastry, and Y.~Ma, ``Robust face
  recognition via sparse representation,'' \emph{IEEE Transactions on Pattern
  Analysis and Machine Intelligence}, vol.~31, no.~2, pp. 210--227, 2009.

\bibitem{mckaymongaraj17}
J.~McKay, V.~Monga, and R.~G. Raj, ``Robust sonar atr through bayesian
  pose-corrected sparse classification,'' \emph{IEEE Transactions on Geoscience
  and Remote Sensing}, vol.~55, no.~10, pp. 5563--5576, 2017.

\bibitem{MCA}
J.-L. Starck, Y.~Moudden, J.~Bobin, M.~Elad, and D.~Donoho, ``Morphological
  component analysis,'' \emph{Proceedings of SPIE - The International Society
  for Optical Engineering}, vol. 5914, 08 2005.

\bibitem{COLBROOK2024127}
\BIBentryALTinterwordspacing
M.~J. Colbrook, ``Chapter 4 - the multiverse of dynamic mode decomposition
  algorithms,'' in \emph{Numerical Analysis Meets Machine Learning}, ser.
  Handbook of Numerical Analysis, S.~Mishra and A.~Townsend, Eds.\hskip 1em
  plus 0.5em minus 0.4em\relax Elsevier, 2024, vol.~25, pp. 127--230. [Online].
  Available:
  \url{https://www.sciencedirect.com/science/article/pii/S1570865924000048}
\BIBentrySTDinterwordspacing

\bibitem{grosek2014dynamicmodedecompositionrealtime}
\BIBentryALTinterwordspacing
J.~Grosek and J.~N. Kutz, ``Dynamic mode decomposition for real-time
  background/foreground separation in video,'' 2014. [Online]. Available:
  \url{https://arxiv.org/abs/1404.7592}
\BIBentrySTDinterwordspacing

\bibitem{kamilov2017plug}
U.~S. Kamilov, H.~Mansour, and B.~Wohlberg, ``A plug-and-play priors approach
  for solving nonlinear imaging inverse problems,'' \emph{IEEE Signal
  Processing Letters}, vol.~24, no.~12, pp. 1872--1876, 2017.

\bibitem{croitoru2023diffusion}
F.-A. Croitoru, V.~Hondru, R.~T. Ionescu, and M.~Shah, ``Diffusion models in
  vision: A survey,'' \emph{IEEE transactions on pattern analysis and machine
  intelligence}, vol.~45, no.~9, pp. 10\,850--10\,869, 2023.

\bibitem{su2024adobe240fps}
S.~Su, M.~Delbracio, J.~Wang, G.~Sapiro, W.~Heidrich, and O.~Wang, ``{Dataset:
  Adobe 240-fps},'' \url{https://doi.org/10.57702/zod0h8e9}, 2024, accessed:
  2025-05-05.

\bibitem{H_Tu_2014}
\BIBentryALTinterwordspacing
J.~H.~Tu, C.~W.~Rowley, D.~M.~Luchtenburg, S.~L.~Brunton, and J.~Nathan~Kutz,
  ``On dynamic mode decomposition: Theory and applications,'' \emph{Journal of
  Computational Dynamics}, vol.~1, no.~2, p. 391–421, 2014. [Online].
  Available: \url{http://dx.doi.org/10.3934/jcd.2014.1.391}
\BIBentrySTDinterwordspacing

\bibitem{Brunton_Kutz_2019}
S.~L. Brunton and J.~N. Kutz, \emph{Data-Driven Science and Engineering:
  Machine Learning, Dynamical Systems, and Control}.\hskip 1em plus 0.5em minus
  0.4em\relax Cambridge University Press, 2019.

\bibitem{ROWLEY_MEZIC_BAGHERI_SCHLATTER_HENNINGSON_2009}
C.~W. ROWLEY, I.~MEZIĆ, S.~BAGHERI, P.~SCHLATTER, and D.~S. HENNINGSON,
  ``Spectral analysis of nonlinear flows,'' \emph{Journal of Fluid Mechanics},
  vol. 641, p. 115–127, 2009.

\bibitem{doi:10.1073/pnas.1517384113}
\BIBentryALTinterwordspacing
S.~L. Brunton, J.~L. Proctor, and J.~N. Kutz, ``Discovering governing equations
  from data by sparse identification of nonlinear dynamical systems,''
  \emph{Proceedings of the National Academy of Sciences}, vol. 113, no.~15, pp.
  3932--3937, 2016. [Online]. Available:
  \url{https://www.pnas.org/doi/abs/10.1073/pnas.1517384113}
\BIBentrySTDinterwordspacing

\bibitem{kutz2015multiresolutiondynamicmodedecomposition}
\BIBentryALTinterwordspacing
J.~N. Kutz, X.~Fu, and S.~L. Brunton, ``Multi-resolution dynamic mode
  decomposition,'' 2015. [Online]. Available:
  \url{https://arxiv.org/abs/1506.00564}
\BIBentrySTDinterwordspacing

\bibitem{Dylewsky_2019}
\BIBentryALTinterwordspacing
D.~Dylewsky, M.~Tao, and J.~N. Kutz, ``Dynamic mode decomposition for
  multiscale nonlinear physics,'' \emph{Physical Review E}, vol.~99, no.~6,
  Jun. 2019. [Online]. Available:
  \url{http://dx.doi.org/10.1103/PhysRevE.99.063311}
\BIBentrySTDinterwordspacing

\bibitem{Starck2005ImageDecomposition}
J.-L. Starck, M.~Elad, and D.~L. Donoho, ``Image decomposition via the
  combination of sparse representations and a variational approach,''
  \emph{IEEE Transactions on Image Processing}, vol.~14, no.~10, pp.
  1570--1582, 2005.

\bibitem{Elad2005MCAInpainting}
M.~Elad, J.-L. Starck, D.~L. Donoho, and P.~Querre, ``Simultaneous cartoon and
  texture image inpainting using morphological component analysis (mca),''
  \emph{Applied and Computational Harmonic Analysis}, vol.~19, no.~3, pp.
  340--358, 2005.

\bibitem{Starck2004RedundantTransforms}
J.-L. Starck, M.~Elad, and D.~L. Donoho, ``Redundant multiscale transforms and
  their application for morphological component analysis,'' \emph{Advances in
  Imaging and Electron Physics}, vol. 132, pp. 287--348, 2004.

\bibitem{BSS}
M.~Pal, R.~Roy, J.~Basu, and M.~S. Bepari, ``Blind source separation: A review
  and analysis,'' in \emph{2013 International Conference Oriental COCOSDA held
  jointly with 2013 Conference on Asian Spoken Language Research and Evaluation
  (O-COCOSDA/CASLRE)}, 2013, pp. 1--5.

\bibitem{NRL_2017}
M.~Farshchian, ``Target extraction and imaging of maritime targets in the sea
  clutter spectrum using sparse separation,'' \emph{IEEE Geoscience and Remote
  Sensing Letters}, vol.~14, no.~2, pp. 232--236, 2017.

\bibitem{RIZAEV2022120}
\BIBentryALTinterwordspacing
I.~G. Rizaev, O.~Karakuş, S.~J. Hogan, and A.~Achim, ``Modeling and sar
  imaging of the sea surface: A review of the state-of-the-art with
  simulations,'' \emph{ISPRS Journal of Photogrammetry and Remote Sensing},
  vol. 187, pp. 120--140, 2022. [Online]. Available:
  \url{https://www.sciencedirect.com/science/article/pii/S0924271622000594}
\BIBentrySTDinterwordspacing

\bibitem{Maggioni2012}
M.~Maggioni, G.~Boracchi, A.~Foi, and K.~Egiazarian, ``Video denoising,
  deblocking, and enhancement through separable 4-d nonlocal spatiotemporal
  transforms,'' \emph{IEEE Transactions on Image Processing}, vol.~21, no.~9,
  pp. 3952--3966, September 2012.

\bibitem{ISAR_book}
V.~Chen and M.~Martorella, \emph{Inverse Synthetic Aperture Radar Imaging:
  Principles, Algorithms and Applications}, 09 2014.

\bibitem{hjw9-d428-21}
\BIBentryALTinterwordspacing
N.~Pandey and S.~Sundar~Ram, ``Dataset of simulated inverse synthetic aperture
  radar (isar) images of automotive targets.'' 2021. [Online]. Available:
  \url{https://dx.doi.org/10.21227/hjw9-d428}
\BIBentrySTDinterwordspacing

\bibitem{9695280}
N.~Pandey and S.~S. Ram, ``Classification of automotive targets using inverse
  synthetic aperture radar images,'' \emph{IEEE Transactions on Intelligent
  Vehicles}, vol.~7, no.~3, pp. 675--689, 2022.

\end{thebibliography}

	\vfill

\end{document}